\documentclass[conference]{IEEEtran}
\IEEEoverridecommandlockouts
\usepackage{cite}
\usepackage{amsmath,amssymb,amsfonts}
\usepackage{bbm}
\usepackage{algorithmic}
\usepackage{graphicx}
\usepackage{textcomp}
\usepackage{xcolor}

\usepackage{color}

\def\BibTeX{{\rm B\kern-.05em{\sc i\kern-.025em b}\kern-.08em
    T\kern-.1667em\lower.7ex\hbox{E}\kern-.125emX}}
\begin{document}

\title{Weighted Point Cloud Normal Estimation\\
\thanks{$^*$ Xuequan Lu is the corresponding author. }
}

\author{
\IEEEauthorblockN{Weijia Wang$^a$, Xuequan Lu$^{a*}$, Di Shao$^a$, Xiao Liu$^a$, Richard Dazeley$^a$, Antonio Robles-Kelly$^a$, Wei Pan$^b$}
\IEEEauthorblockA{$^a$ \textit{Deakin University, Australia}}
\IEEEauthorblockA{$^b$ \textit{OPT Machine Vision Tech Co., Ltd, Japan}}
\IEEEauthorblockA{\{wangweijia, xuequan.lu, shaod, xiao.liu, richard.dazeley, antonio.robles-kelly\}@deakin.edu.au, vpan@foxmail.com}

}

\maketitle

\begin{abstract}
  Existing normal estimation methods for point clouds are often less robust to severe noise and complex geometric structures. Also, they usually ignore the contributions of different neighbouring points during normal estimation, which leads to less accurate results. In this paper, we introduce a weighted normal estimation method for 3D point cloud data. We innovate in two key points: 1) we develop a novel weighted normal regression technique that predicts point-wise weights from local point patches and use them for robust, feature-preserving normal regression; 2) we propose to conduct contrastive learning between point patches and the corresponding ground-truth normals of the patches' central points as a pre-training process to facilitate normal regression. Comprehensive experiments demonstrate that our method can robustly handle noisy and complex point clouds, achieving state-of-the-art performance on both synthetic and real-world datasets. 
\end{abstract}

\begin{IEEEkeywords}
3D Point Cloud, Normal Estimation
\end{IEEEkeywords}

\section{Introduction}
\label{sec:intro}

Point clouds are used in a vast range of fields, such as robotics, autonomous driving, 3D scanning and modelling. However, raw point cloud data coming from sensing devices is unordered and does not equip with normal information. Also, it is often corrupted with noise due to precision limitation of sensing devices. As a remedy solution, estimating accurate normals for point clouds has been proven effective in enhancing the performance in various tasks, such as noise filtering~\cite{lu_lowrank_2018, Lu_pointfilter_2021,Lu_DFP_2020,Lu_GPF_2018} and surface reconstruction~\cite{kazhdan_poisson_2006}.

Conventional normal estimation methods such as Principal Component Analysis (PCA)~\cite{Hoppe_1992} show limited capability in handling noise or complex point cloud surfaces. To handle such defects, learning-based normal regression methods have been proposed. For example, PCPNet~\cite{Guerrero_pcpnet_2018} and Nesti-Net~\cite{Benshabat_Nesti_2019} regress normals from the encoded features of local neighbourhoods (as local patches) on point clouds. Although they show improved robustness to noise, they ignore the varying contributions from the neighbouring points within local patches, which may lead to inaccurate results on complex surfaces. In recent years, methods leveraging surface fitting~\cite{Shabat_deepfit_2020, Zhu_adafit_2021} have been introduced, which attempt to fit polynomial surfaces onto point clouds in order to approximate normals. While such methods demonstrate adaptiveness to complicated geometric surfaces, they tend to be less robust to point clouds contaminated by severe noise. 

\begin{figure}[!t]
\centering{
	\includegraphics[width=\columnwidth]{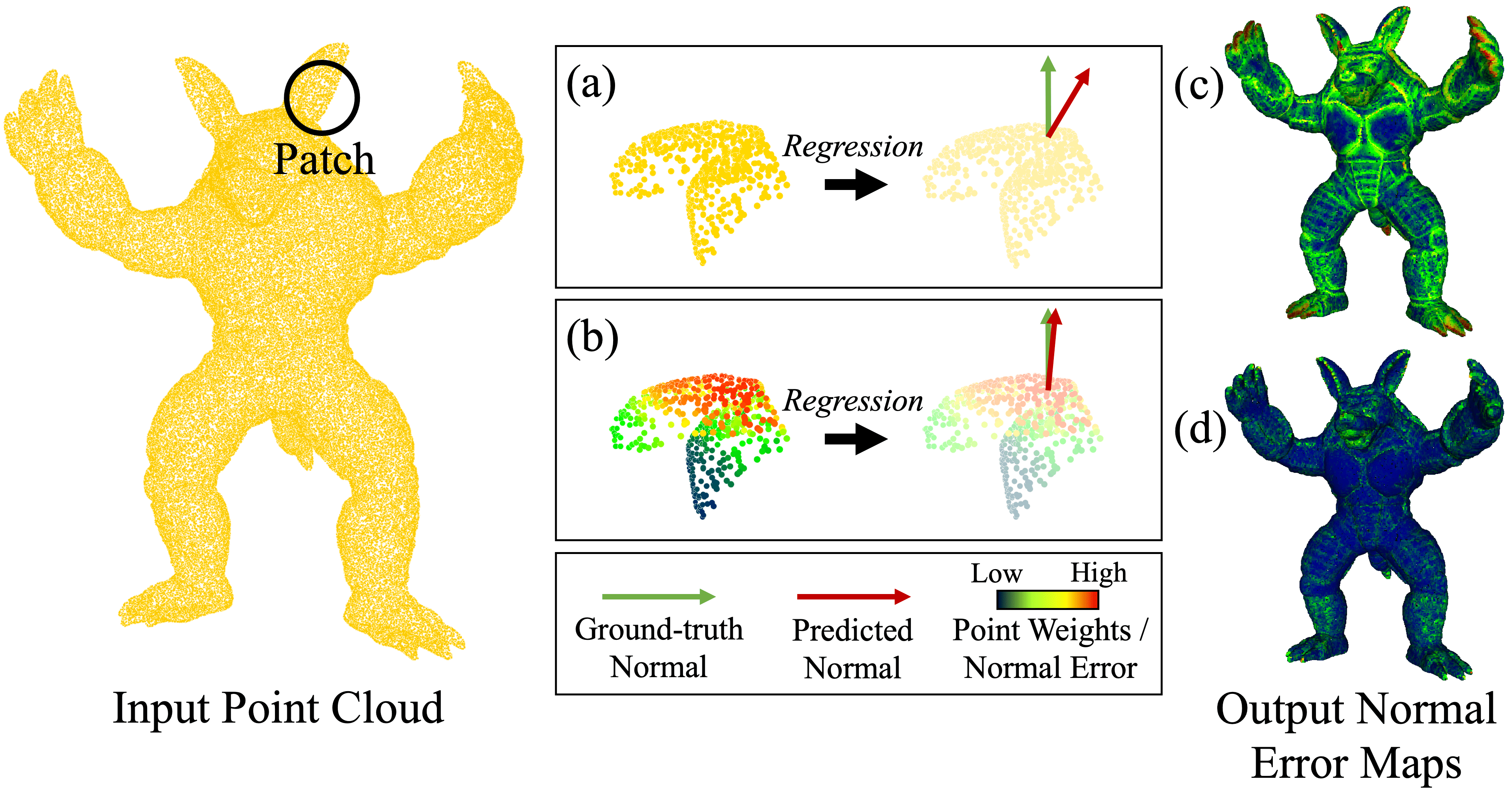}
	}
\caption{For any input point cloud patch, regressing the normal with uniform weights on all points (shown in (a)) leads to less accurate results (shown in (c)). By contrast, estimating point-wise weights and performing weighted normal regression (shown in (b)) result in improved normal prediction accuracy (shown in (d)). }
\label{fig:teaser}
\end{figure}

In this paper, we are motivated to exploit the relevance of neighbouring points to the patch's central point during normal estimation. In specific, we aim to model point contributions as weights and utilise them for normal regression in order to reduce the negative impacts from less relevant points and improve the accuracy of normal prediction. We therefore propose a novel weighted normal regression method for point clouds that consists of a \textit{contrastive pre-training} stage and a \textit{normal estimation} stage. In our pre-training stage, we propose to leverage the correspondences between each normal and the nearby relevant points in a contrastive learning manner. To realise this, we train a point encoder associated with a weight regressor to output weighted point features within a patch, and contrast them with the feature of the central point's normal. In our normal estimation training stage, we fine-tune the pre-trained point encoder and weight regressor, and feed the output weighted point features into our normal regressor to estimate the normal for each patch's central point. By doing so, our method puts higher weights on more relevant points and leads to feature-preserving and noise-resisting results. Fig.~\ref{fig:teaser} provides an illustration of the uniformly-weighted scheme and our approach. 

Our key contributions are summarised as follows:
\begin{itemize}
    \item We propose a novel weighted normal regression method for point clouds, which effectively preserves geometric features and is robust to noise.  
    \item We introduce a weight regressor to predict point-wise weights from local patches and use them for more accurate point feature learning and normal regression. 
    \item We propose to contrast patch points and the corresponding normal of the patch's central point in order to exploit the relations between them and facilitate normal regression. 
\end{itemize}

\section{Related Work}
\label{sec:related-work}
\subsection{Point Cloud Normal Estimation}
Normal estimation for point clouds is a fundamental research problem in the 3D field. A typical example is based on PCA~\cite{Hoppe_1992}, a method that calculates each point's normal by computing eigenvectors from the covariance matrix based on the neighbouring points. While it is simple and straightforward, it tends to blur the details and is fragile on noisy point clouds. With the development in deep learning, data-driven learning-based approaches start to step into researchers' attention, demonstrating improved performance on normal estimation compared with conventional methods. An example is PCPNet~\cite{Guerrero_pcpnet_2018}, which utilises convolutional neural networks to extract features from local point patches and regresses normals from them. To further increase the prediction accuracy, Nesti-Net~\cite{Benshabat_Nesti_2019} utilises a mixture-of-experts network to encode patches of various scales and regress normals from the optimal ones. While such regression-based methods are robust to noise, they omit the different contributions from neighbouring points to the patch's central point during normal estimation. In recent years, there are methods attempting to fit surfaces on point clouds to approximate normals, which take point contributions into consideration. For instance, Lenssen et al.~\cite{Lenssen_DI_2020} introduced Deep Iterative (DI), which adopts graph neural networks to perform weighted least squares plane fitting on point neighbourhoods. Similarly, DeepFit~\cite{Shabat_deepfit_2020} attempts to fit polynomial surfaces on point clouds to approximate normals for them. Based on DeepFit, AdaFit~\cite{Zhu_adafit_2021} additionally predicts point-wise offsets and uses a cascaded scale aggregation module to enhance the performance of surface fitting on various shapes. Nonetheless, while such methods perform well on clean points, their performance becomes less robust to noisy point clouds, especially for those corrupted with severe noise. 

\subsection{Contrastive Learning}

Contrastive learning has been gaining increasing popularity in both 2D and 3D machine vision tasks over the years. For instance, NPID~\cite{Wu_2018} exploits a discrete memory bank to store instance features to facilitate effective contrastive learning for images, and the idea is further leveraged in~\cite{Zhuang_2019}. Another example is SimCLR~\cite{chen_simclr_2020}, which contrasts pairs of different augmented images (e.g., cropped, flipped or masked) and achieves improved image classification results. For 3D point clouds, contrastive learning also demonstrates its ability to facilitate classification, segmentation and object detection tasks~\cite{xie_pointcontrast_2020,jiang_segmentation_2021}. Extending from learning the same type of data, contrastive learning further shows its strength on learning multi-modal correspondences. For instance, CrossPoint~\cite{afham_crosspoint_2022} contrasts 3D point cloud data and 2D images in order to find potential correspondences which facilitate point cloud understanding tasks. Nevertheless, there has been little attention on extending contrastive learning to point cloud normal estimation. In particular, given that contrastive learning has the ability to find out cross-modal correspondences, we are motivated to directly contrast points and normals in order to facilitate normal estimation.

\begin{figure*}[!ht]
	\centering{
	\includegraphics[width=\textwidth]{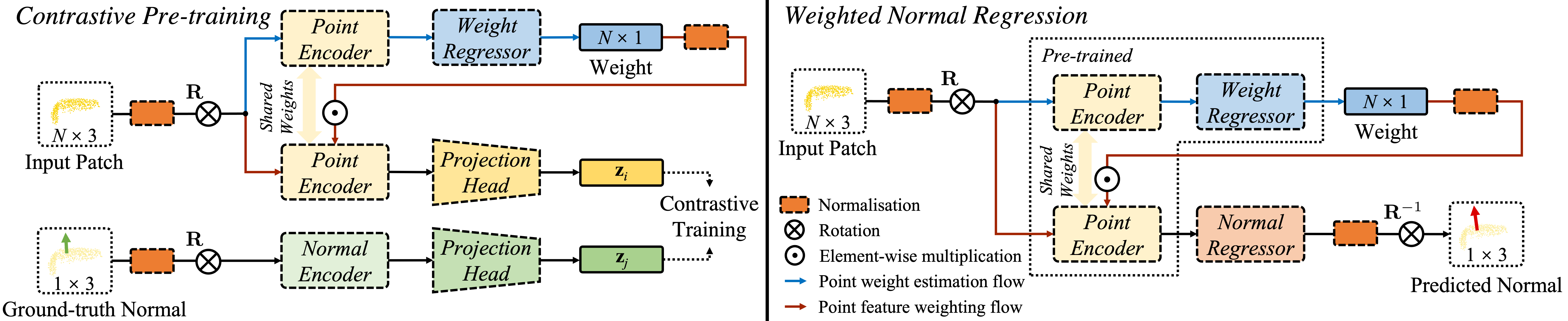}
	}
	\caption{Overview of our method. In our contrastive pre-training stage (left), we train our network in a dual-branch contrastive learning manner. In our weighted normal regression stage (right), we train the normal regressor together with the pre-trained point encoder and weight regressor to estimate normals.} 
	\label{fig:pipeline}
\end{figure*}

\section{Method}
\label{sec:method}

\subsection{Overview}

Given a point cloud $\mathbf{P} = \{\mathbf{p}_1,\ldots, \mathbf{p}_m\ |\ \mathbf{p}_i \in \mathbb{R}^3,\ i\ = 1, ..., m \}$, we aim to predict the normal ${\mathbf{n}_i}'$ of each point $\mathbf{p}_i$ from the features of patch $\mathcal{P}_i$, which centers at $\mathbf{p}_i$ and contains $N$ nearest neighbouring points. It is worth noting that the raw patch $\mathcal{P}_i$ may be of any size or with arbitrary degrees of freedom and is thus unsuitable for direct training. To alleviate this issue, we normalise $\mathcal{P}_i$ to a unit sphere and align it with the global space using a rotation matrix $\mathbf{R}$, which is computed by PCA decomposition on $\mathcal{P}_i$. We later normalise the predicted normal's length and map it back to its original orientation using the inverse matrix $\mathbf{R}^{-1}$ during testing. 

Our overall pipeline is shown in Fig.~\ref{fig:pipeline}. We first pre-train our point encoder and weight regressor using a dual-branch contrastive learning technique, which is elaborated in Sec.~\ref{sec:contrastive-learning}. We then fine-tune the trained point encoder and weight regressor to output weighted patch features for downstream normal regression, which is explained in Sec.~\ref{sec:normal-est}.

\subsection{Contrastive Pre-training}
\label{sec:contrastive-learning}
By doing contrastive pre-training, we aim to exploit the correspondences between each patch and the normal of its central point (as a patch-normal pair). As there are two types of input data, only using a single encoder is not feasible in this case. 
Also, as not all points within a patch are highly relevant to the central point's normal, we need to reduce the weights from less relevant points. We thus develop a dual-branch network, where the first branch consists of a point encoder associated with a point weight regressor, and the other branch contains a normal encoder. In addition, each branch is equipped with a projection head. We explain the details as follows.

\textit{Point encoder.} The point encoders in prior normal estimation work widely adopt point convolutional networks such as~\cite{Qi_2017}. However, such encoders only consider point-wise information and ignore the relationships between each point and its neighbours. Previous literature~\cite{wang_dgcnn_2019} has proven that dynamically constructing graphs based on each point's local neighbourhood is effective in revealing latent geometric relationships among the neighbours. We are motivated that such relationships contain useful patch features that can enhance normal estimation, especially on noisy and complex surfaces. Thus, we adopt a neighbourhood-based graph convolutional network to encode our patches. In specific, we utilise the EdgeConv modules from DGCNN~\cite{wang_dgcnn_2019} as the encoding layers for our point encoder. For each point (or vertex) $\mathbf{x}_i$ in the input data, the output feature ${\mathbf{x}_i}'$ from each encoding layer is defined as
\begin{equation} 
\label{eq:edgeconv-layer}
    {{\mathbf{x}_i}'} = max_{j: (i, j) \in \mathcal{E}}{f(\mathbf{x}_i, \mathbf{x}_j - \mathbf{x}_i)},
\end{equation}
where $\mathcal{E}$ is the set of edges in the local graph formed by central vertex $\mathbf{x}_i$ and its $k$ nearest neighbouring vertices, and $\mathbf{x}_j$ is a neighbouring vertex within the local graph. We concatenate each edge $(\mathbf{x}_j - \mathbf{x}_i)$ and vertex $\mathbf{x}_i$, and send them to a non-linear function $f$ with learnable parameters followed by a max-pooling operation $max$. We set $k=20$ empirically, and denote the final point-wise features of the patch by $\mathbf{F}_{i}$, which is an $N \times 1024$ matrix. Finally, we apply pooling operations to obtain the global permutation-invariant features for the patch.

\textit{Weight regressor.} After obtaining the point patch features, we feed them into our weight regressor, which comprises multi-layer perceptrons (MLPs), to predict the relevance of each point to the patch's central point as weights. To train the weight regressor, we utilise the ground-truth normals of the neighbouring points within the patch, as normals are direct representations of the underlying surfaces. A neighbour point is considered as highly relevant if its normal and the central point's normal have a high cosine similarity value, and is less relevant otherwise. Based on this, we compute the squared cosine values between the central point's normal and each neighbour's normal within the patch as our ground-truth weights. \textit{Note that directly utilising such ground-truth weights for normal prediction is not feasible, as they are not available during testing.} We denote the predicted point-wise weights for patch $\mathcal{P}_i$ as $\mathbf{W}_i = (w_j\ |\ j=1,\ldots, N)$ and thus formulate the weight regression loss as
\begin{equation} 
\label{eq:weight-loss}
    L_{weight} = \sum_{j=1}^{N}(w_{j} - ({\mathbf{n}_i \cdot \mathbf{n}_j})^2)^2,
\end{equation}
where $\mathbf{n}_i$ is the ground-truth normal of the central point $\mathbf{p}_i$, and $\mathbf{n}_j$ is the ground-truth normal of each point $\mathbf{p}_j$ within patch $\mathcal{P}_i$. Once we obtain the predicted weights, we perform element-wise multiplication for $\mathbf{W}_i$ and $\mathbf{F}_{i}$ and obtain the weighted point-wise patch feature matrix ${\mathbf{F}_{i}}'$ (an $N \times 1024$ matrix), with the process being denoted as ${\mathbf{F}_{i}}' = \mathbf{W}_i \odot \mathbf{F}_{i}$.
Finally, we apply pooling operations on ${\mathbf{F}_{i}}'$ to obtain the global permutation-invariant feature for each patch.

\textit{Normal encoder.} The normal of the central point of each patch is a $1 \times 3$ vector and thus cannot be fed into the aforementioned graph convolutional network. We therefore encode the normal into a $1 \times 1024$ feature vector using MLPs. 

\textit{Projection heads.} The raw point and normal feature vectors are in a high-dimensional space with plenty of redundant information that intervenes effective contrastive learning. We thus feed each feature vector respectively into a projection head and project them to a lower-dimensional space, which is a remedy solution to reduce redundancy~\cite{chen_simclr_2020}. Here, we denote each projected point patch feature vector as $\mathbf{z}_i$ and the projected normal feature vector as $\mathbf{z}_j$. 

We then formulate our loss function for contrastive training. Following~\cite{chen_simclr_2020}, we define a temperature-scaled cross entropy loss $l_{i, j}$ for each patch-normal pair as
\begin{equation} 
\label{eq:contrastive-l-pair-term}
    l_{i, j} = -log\frac{\exp(sim(\mathbf{z}_{i}, \mathbf{z}_{j})/\tau)}{\sum_{k=1}^{2B}\mathbbm{1}_{k\neq i}\exp(sim(\mathbf{z}_{i}, \mathbf{z}_{k})/\tau)},
\end{equation}
where $sim(\mathbf{z}_{i}, \mathbf{z}_{j})$ is the cosine similarity between $\mathbf{z}_{i}$ and $\mathbf{z}_{j}$, and is divided by a temperature parameter $\tau$. $\mathbbm{1}_{k\neq i} \in \{0, 1\}$ is an indicator function that becomes 1 if and only if $k\neq i$. 
For each patch-normal pair, we compute the temperature-scaled cross entropy loss between $(\mathbf{z}_{i},\mathbf{z}_{j})$ and $(\mathbf{z}_{j},\mathbf{z}_{i})$; we then sum up the losses for all pairs within the $B$-sized batch and take the average loss as our overall contrastive pre-training loss $L_{cont}$. 
By doing so, the point encoder can maximise the correspondences within each patch-normal pair and minimise the similarities among different pairs. When pre-training finishes, we keep the trained point encoder and weight regressor for fine-tuning in the subsequent normal estimation training process. 

Combining $L_{cont}$ and $L_{weight}$, the overall loss of our contrastive pre-training stage $L_{pre}$ is defined as
\begin{equation} 
\label{eq:weighted-pretrain-loss}
    L_{pre} = L_{cont} + \alpha L_{weight},
\end{equation}
where $\alpha$ is a hyper-parameter controlling the ratio of $L_{weight}$ and we set it to 1.0 based on experiment results.

\subsection{Normal Estimation}
\label{sec:normal-est}
We fine-tune our pre-trained point encoder and weight regressor in our normal estimation stage. As shown in Fig.~\ref{fig:pipeline}, the output patch features from our point encoder and weight regressor are fed into our normal regression module, which consists of a series of fully connected layers as MLPs. The normal regression module produces a $1 \times 3$ vector as the predicted normal for each input patch's central point $\mathbf{p}_i$. To minimise the angle between each predicted normal and the ground-truth one, we design a cosine similarity loss function $L_{cos}$ as a basic loss term which is defined as
\begin{equation} 
\label{eq:cosine-loss}
    L_{cos} = 1 - ({\mathbf{n}_{i}}' \cdot \mathbf{n}_i)^2,
\end{equation}
where ${\mathbf{n}_{i}}'$ is the predicted normal, and $\mathbf{n}_i$ is the ground-truth normal of the patch's central point $\mathbf{p}_i$. Nonetheless, merely utilising the cosine loss is not enough as it treats all points equally and omits the different contributions from the neighbouring points. To alleviate this issue, we introduce $L_{weight}$ from Eq.~\eqref{eq:weight-loss} to fine-tune our network. The loss for our downstream normal estimation stage $L_{down}$ is therefore defined as 
\begin{equation} 
\label{eq:down-total-loss}
    L_{down} = L_{cos} + \alpha L_{weight},
\end{equation}
where we set $\alpha$ to 1.0 as per Eq.~\eqref{eq:weighted-pretrain-loss} for consistency during training. We display our experimental results on different network configurations and loss terms in Sec.~\ref{sec:ablation}. \textit{More details of our network architecture are provided in the supplementary material.}


\begin{table*}[t]
\centering
\caption{Quantitative evaluation (RMSE) on shapes with Gaussian noise from PCPNet dataset~\cite{Guerrero_pcpnet_2018}. For each method, we compute the average RMSE of all shapes at different noise levels and show the results in degrees. }
\label{tab:pcpnet-result}
\begin{tabular}{|c|ccccccccc|}
\hline
Aug. Noise Level  & Ours           & AdaFit & DeepFit        & DI    & Nesti-Net & PCPNet & PCA (small) & PCA (medium) & PCA (large) \\ \hline
0.36\%            & 13.25          & 13.39  & \textbf{13.18} & 14.00 & 15.10     & 15.92  & 29.49     & 15.07     & 17.54     \\
0.60\%            & \textbf{16.38} & 16.44  & 16.72          & 17.18 & 17.63     & 18.26  & 41.82     & 18.47     & 18.99     \\
0.84\%              & \textbf{18.57} & 18.85  & 19.55          & 19.37 & 19.64     & 20.21  & 48.40     & 22.27     & 20.87     \\
1.20\%            & \textbf{21.48} & 21.94  & 23.12          & 21.96 & 22.28     & 22.80  & 53.34     & 27.72     & 23.54     \\ \hline
Avg. RMSE & \textbf{17.42} & 17.66  & 18.14          & 18.12 & 18.66     & 19.30  & 43.26     & 20.88     & 20.23     \\ \hline
\end{tabular}
\end{table*}

\section{Experiments and Results}
\label{sec:exps-and-results}

\subsection{Training and Implementation Details}
We adopt the same dataset as PCPNet~\cite{Guerrero_pcpnet_2018}, including the train-test split and data augmentation settings (i.e., adding noise). We implement our framework using PyTorch, and train and test it on an NVIDIA GeForce RTX 3080 10GB GPU. We pre-train our model for 50 epochs with two Adam optimisers respectively for each encoder, and then train 100 epochs for normal estimation with a single Adam optimiser. In both training stages, we set the learning rate to 0.001 with a batch size of 32. For each point patch $\mathcal{P}_i$, we set the number of points $N$ to 700 during our experiments.

\subsection{Comparison Metrics and Methods}
Following prior work, we use Root Mean Square Error (RMSE) to measure the angular accuracy for our predicted normals. We compare against the conventional PCA method~\cite{Hoppe_1992}, surface fitting methods including DI~\cite{Lenssen_DI_2020}, DeepFit~\cite{Shabat_deepfit_2020} and AdaFit~\cite{Zhu_adafit_2021}, as well as regression-based methods including PCPNet~\cite{Guerrero_pcpnet_2018} and Nesti-Net~\cite{Benshabat_Nesti_2019}. For the PCA method, we set 3 scales of $k$-nearest neighbours as per~\cite{Guerrero_pcpnet_2018}, where $k = 18$, $112$ and $450$ respectively for small, medium and large patches. 


\subsection{Quantitative Evaluation Results}
\begin{figure}[t]
\centering{
	\includegraphics[width=\columnwidth]{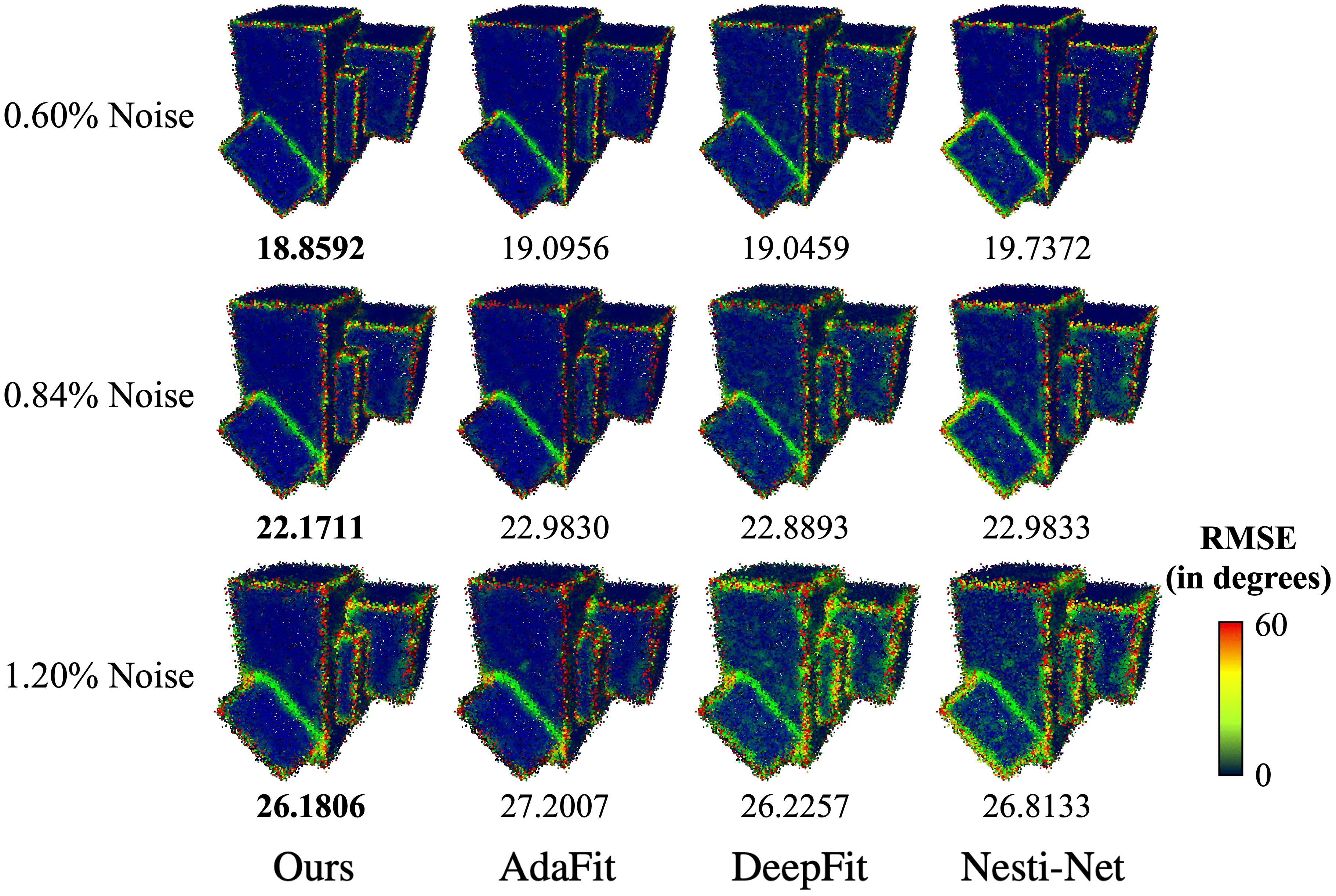}
	}
\caption{Visualised RMSE on Boxunion, where our method achieves the minimum error on all noise levels. }
\label{fig:pcpnet-robustness}
\end{figure}

\textit{Synthetic dataset.} We first demonstrate the performance on synthetic point clouds corrupted with Gaussian noise. The test set involves 19 shapes that come from PCPNet~\cite{Guerrero_pcpnet_2018}, where each shape also has 4 variants of noise levels (i.e., 0.36\%, 0.6\%, 0.84\% and 1.2\% of the length of the shape's bounding box diagonal). Such noisy shapes' original geometric information is contaminated, bringing significant challenges to feature preservation during normal estimation. Despite the challenge, our method still robustly handles the noise and achieves the state-of-the-art performance with regards to RMSE. As shown in Table \ref{tab:pcpnet-result}, our method outperforms all others at 0.6\%, 0.84\% and 1.2\% noise levels and achieves the minimum overall error. We further demonstrate our method's robustness by visualising the RMSE results on the Boxunion shape from PCPNet's dataset (corrupted with 0.6\%, 0.84\% and 1.2\% noise levels), which is shown in Fig.~\ref{fig:pcpnet-robustness}. The results illustrate that our method achieves the minimum error at each noise level compared with the state-of-the-art methods.

\begin{figure}[t]
\centering{
	\includegraphics[width=\columnwidth]{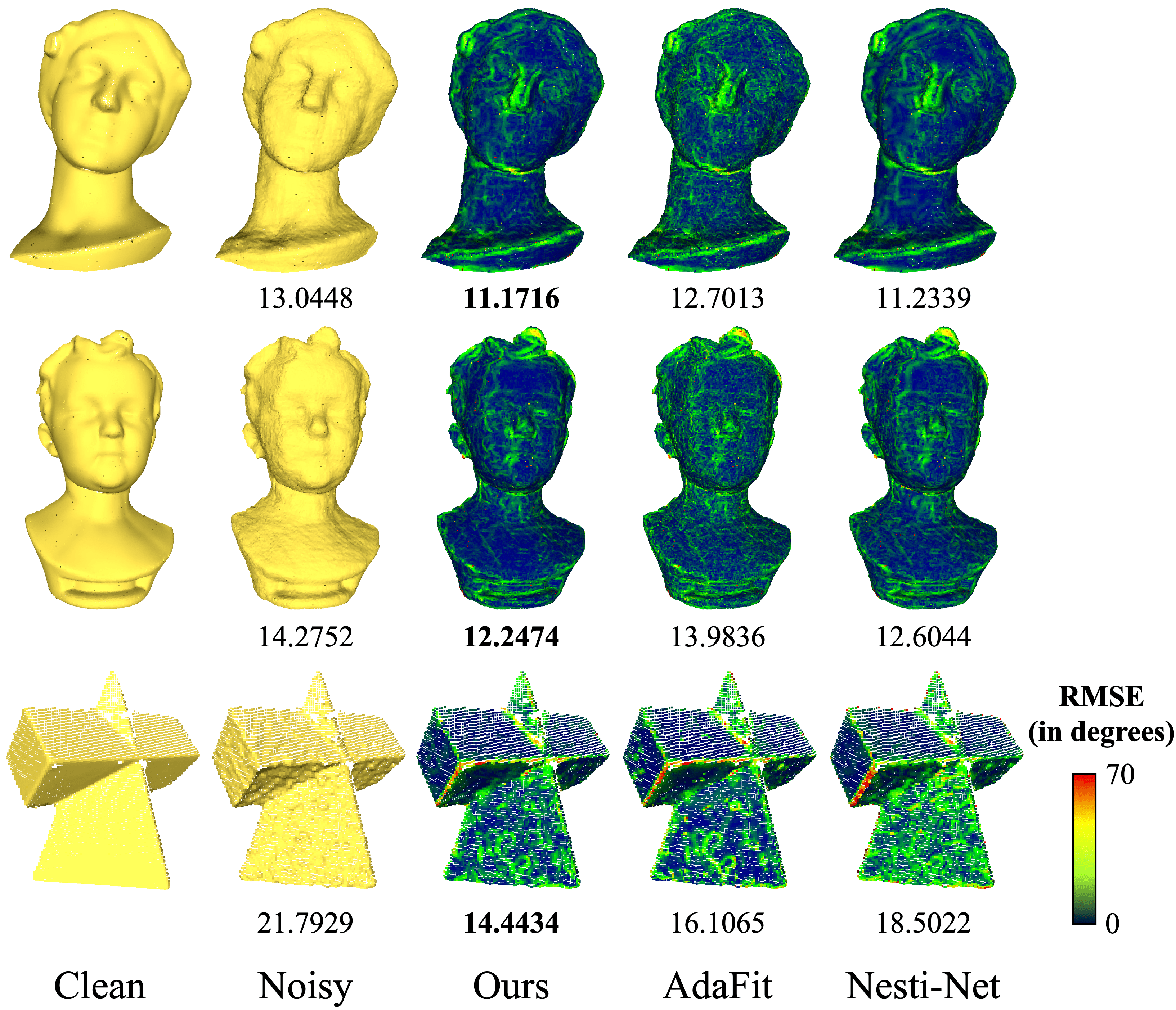}
	}
\caption{Visualised RMSE on Kinect Fusion shapes.}
\label{fig:kinect-fusion}
\end{figure}

\textit{Real-world scanned dataset.} We also display results on raw point clouds from Kinect Fusion~\cite{izadi_kinectfusion_2011} dataset, which are corrupted with noise during scanning. We compare the predicted normals on noisy shapes against the reconstructed clean normals provided by the dataset.
The visualised RMSE results are shown in Fig.~\ref{fig:kinect-fusion}, where both AdaFit and Nesti-Net tend to be sensitive to noise and thus produce inaccurate normals. By contrast, our method performs better on such complex surfaces and achieves lower errors.

\subsection{Visual Results}
\begin{figure}[t]
	\centering{
	
	\includegraphics[width=\columnwidth]{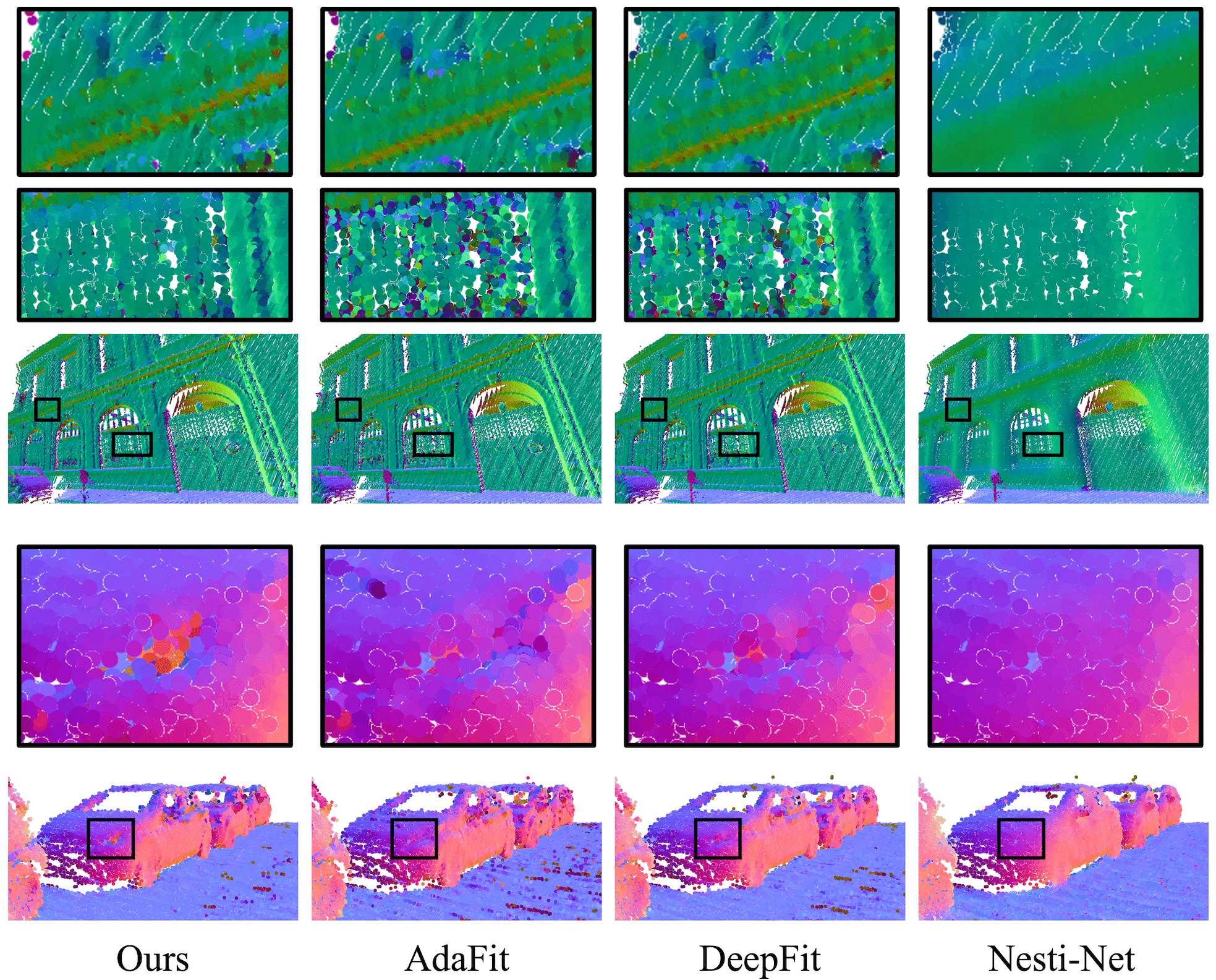}

	}
	\caption{Visual results on Paris-rue-Madame scenes. Colour map is used for better observation on normal orientations. }
	\label{fig:paris}
\end{figure}

We demonstrate visual results on real-world data where the ground-truth is unknown. We adopt the tool by~\cite{huang_ear_2013} to render points and their estimated normals as coloured surfels, where the colours stand for normal orientations, to showcase the normals' quality. Fig.~\ref{fig:paris} demonstrates results on Paris-rue-Madame dataset~\cite{serna-paris-2014} which contains street scene point clouds captured by laser scanners and is contaminated by noise. Although Nesti-Net can produce smoother surfaces, it severely blurs features that should be preserved. The fitting-based methods DeepFit and AdaFit may omit certain small details, as their point convolution-based encoders are less context-aware. In addition, AdaFit's performance becomes less robust on noisy areas such as the road surfaces. By contrast, our method outputs tidier normals on complex areas (such as the window lattice in the first scene) while preserving clear features (such as the building eave in the first scene and the car light in the second scene) compared with others.

	


\subsection{Ablation Study}
\label{sec:ablation}

To evaluate the effects of the components in our proposed method, we demonstrate the average RMSE results on PCPNet's validation set using different network configurations and loss terms, which are shown in Table~\ref{tab:ablation-study}. The configurations include 1) a baseline normal estimation network trained with $L_{cos}$ only, 2) a network pre-trained with $L_{cont}$ and fine-tuned with $L_{cos}$ (i.e., without weight regression), 3) a network with weight regression assisting normal estimation (i.e., without contrastive pre-training), and 4) the full pipeline, where a network is pre-trained and fine-tuned with loss terms in Eq.~\eqref{eq:weighted-pretrain-loss} and Eq.~\eqref{eq:down-total-loss}, respectively. As demonstrated in Table~\ref{tab:ablation-study}, we achieve the best results when we use the full pipeline, i.e., adopt both the contrastive pre-training and the weighted regression strategies.

\textit{Additional experimental results are provided in our supplementary material.} 

\begin{table}[t]
\begin{center}
\caption{Ablation study on different settings. We calculate the average RMSE on PCPNet's validation dataset. *We do not perform contrastive pre-training in this setting.}
\label{tab:ablation-study}
\begin{tabular}{|c|ccc|c|}
\hline

Config. No. & $L_{cos}$ & $L_{cont}$ & $L_{weight}$ & Avg. RMSE \\ \hline
1         & \checkmark      &         &           & 21.15          \\
2         & \checkmark      & \checkmark       &           & 14.70          \\
3*         & \checkmark      &         & \checkmark         & 18.21     \\
4         & \checkmark      & \checkmark       & \checkmark         & \textbf{14.46}     \\ \hline

\end{tabular}
\end{center}
\end{table}

\section{Applications}
\label{sec:applications}
\subsection{Point Cloud Filtering}
\begin{figure}[!t]
\centering{
	\includegraphics[width=\columnwidth]{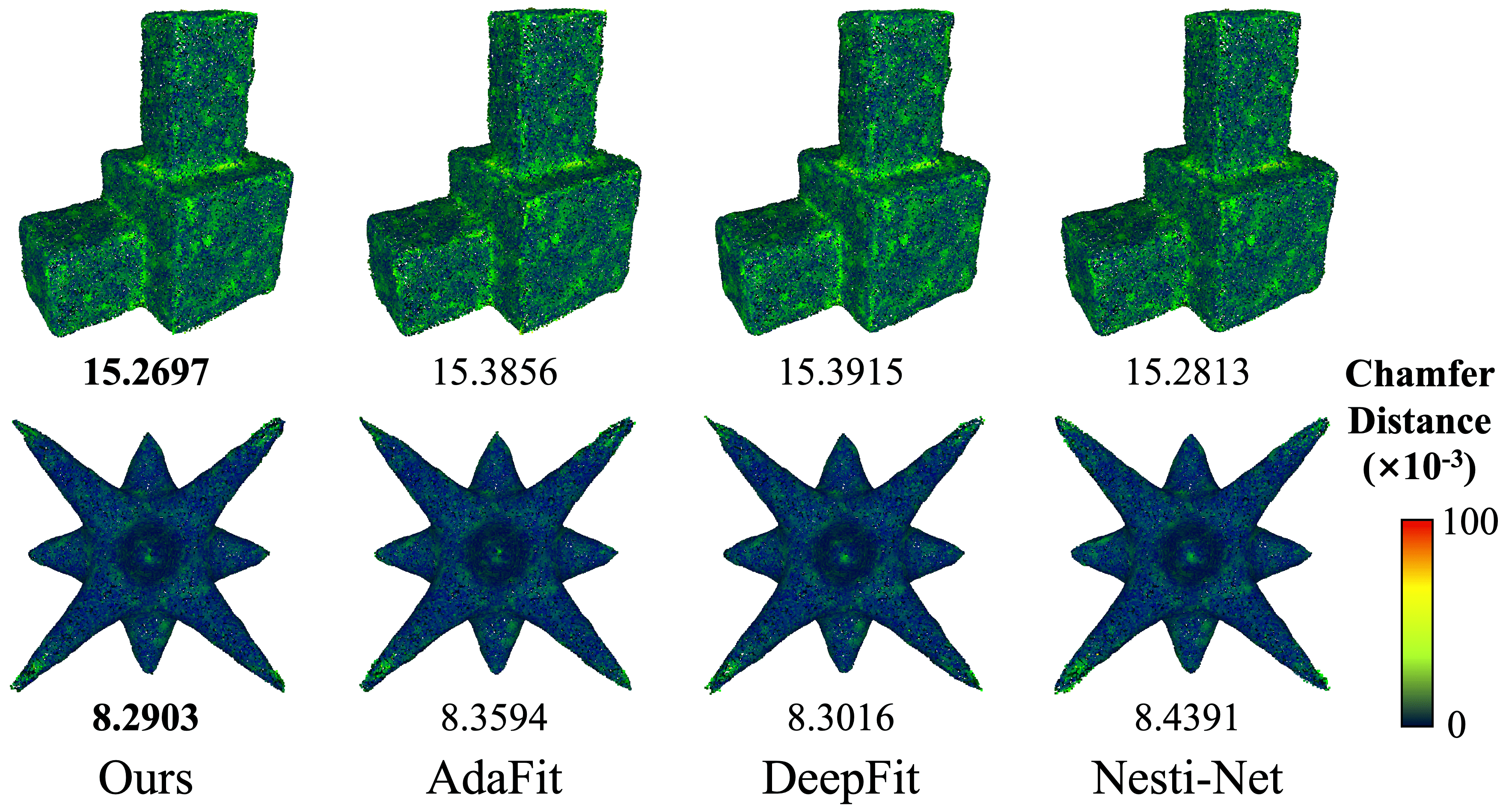}
	}
\caption{Average point-wise Chamfer distance for the filtering results. }
\label{fig:application}
\end{figure}

Point normals can assist with point cloud filtering tasks~\cite{lu_lowrank_2018,Lu_GPF_2018,Lu_DFP_2020}. We adopt the filtering method in~\cite{lu_lowrank_2018} which performs low rank matrix approximation to filter noisy point clouds. This filtering approach relies on input normals, where more accurate normals lead to better filtered results. Here, we use the predicted normals on shapes corrupted with 1.2\% noise from PCPNet's test set, and run filtering for one iteration such that the results solely rely on the input normals. We use Chamfer distance~\cite{open3d_2018} as the metric to measure the filtering quality, which computes the average closest point distances between the filtered points and the ground-truth ones. As shown in Fig.~\ref{fig:application}, the normals predicted by our method lead to the best filtering outcomes.

\subsection{Mesh Surface Reconstruction}
\begin{figure}[!t]
\centering{
	\includegraphics[width=\columnwidth]{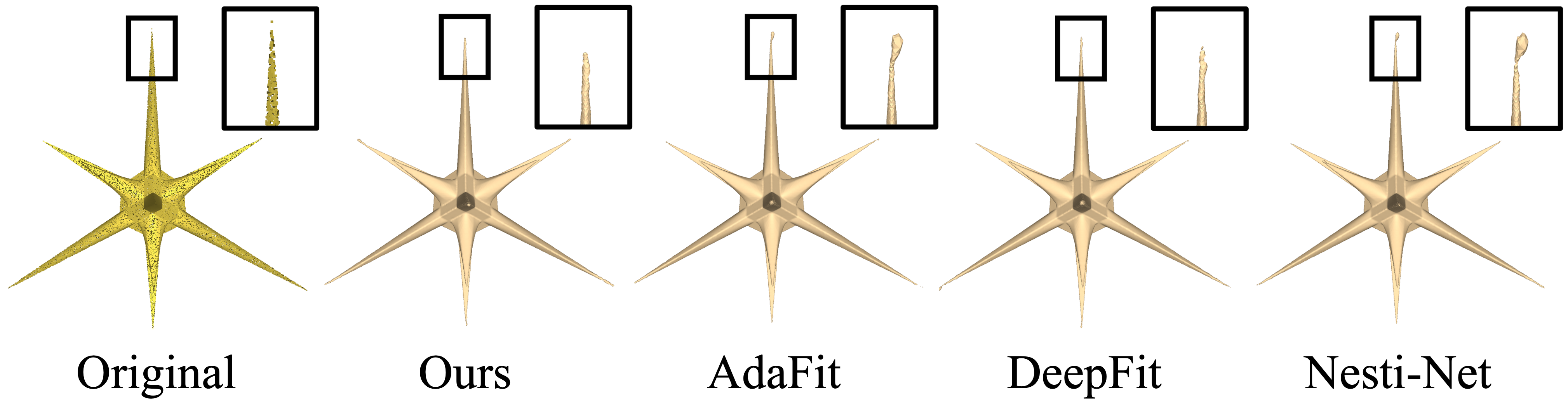}
	}
\caption{Poisson surface reconstruction on a sharp Star shape.}
\label{fig:poisson-visual}
\end{figure}

Accurate normal information can help Poisson mesh reconstruction~\cite{kazhdan_poisson_2006} with reconstructing high-quality mesh surfaces on point clouds. In Fig.~\ref{fig:poisson-visual}, we demonstrate the mesh reconstruction results on a sharp Star shape, which is a tricky case. It shows that the normals predicted by our method lead to more accurate reconstruction on the sharp tip. 

\section{Conclusion}
In this paper, we proposed a novel weighted normal regression method for 3D point clouds. We developed a weight regression module that predicts point-wise weights to apply more weights on relevant points and reduce impacts from less relevant ones. Also, we proposed a contrastive pre-training stage which leverages the relationships between each normal and the nearby relevant points. Extensive experiments demonstrate that our method achieves state-of-the-art performance on both synthetic and real-world datasets, showing robustness to noisy and complex point clouds.

\bibliographystyle{IEEEtran} 
\bibliography{ref2023}


\end{document}